\pdfoutput=1
\documentclass[11pt,a4paper]{article}

\usepackage[]{styles/acl}

\usepackage{times}
\usepackage{latexsym}

\usepackage[T1]{fontenc}

\usepackage[utf8]{inputenc}

\usepackage{microtype}
\usepackage{enumitem}
\usepackage{graphicx}
\usepackage{tabularx}
\usepackage{array}
\usepackage{booktabs}
\usepackage{algpseudocode}
\usepackage{algorithm}
\usepackage{amsmath}
\usepackage[table]{xcolor}

\newcommand\BibTeX{B\textsc{ib}\BibTeX}
\newcommand{\highlight}{\cellcolor{yellow!45}}

\title{Instructional Goal-Aligned Question Generation for Student Evaluation in Virtual Lab Settings: How Closely Do LLMs Actually Align?}

\author{
  \textbf{R. Alexander Knipper}\footnotemark[1]\quad
  \textbf{Indrani Dey}\footnotemark[2]\quad
  \textbf{Souvika Sarkar}\footnotemark[3] \\
  \textbf{Hari Narayanan}\footnotemark[4]\quad
  \textbf{Sadhana Puntambekar}\footnotemark[2]\quad
  \textbf{Santu Karmaker}\footnotemark[1] \\
  Bridge-AI Lab@UCF\footnotemark[1]\quad
  Department of EdPsych, University of Wisconsin-Madison\footnotemark[2] \\
  Department of CS, Wichita State University\footnotemark[3]\quad
  Department of CSSE, Auburn University\footnotemark[4] \\
  \texttt{alexknipper@ucf.edu, idey2@wisc.edu, souvika.sarkar@wichita.edu}\\
  \texttt{naraynh@auburn.edu, puntambekar@education.wisc.edu, santu@ucf.edu} \\}
\date{October 2025}

\begin{document}

\maketitle

\begin{abstract}


Virtual Labs offer valuable opportunities for hands-on, inquiry-based science learning, yet teachers often struggle to adapt them to fit their instructional goals. Third-party materials may not align with classroom needs, and developing custom resources can be time-consuming and difficult to scale. Recent advances in Large Language Models (LLMs) offer a promising avenue for addressing these limitations.
In this paper, we introduce a novel alignment framework for \textit{instructional goal-aligned question generation}, enabling teachers to leverage LLMs to produce simulation-aligned, pedagogically meaningful questions through natural language interaction. The framework integrates four components: instructional goal understanding via teacher–LLM dialogue, lab understanding via knowledge unit and relationship analysis, a question taxonomy for structuring cognitive and pedagogical intent, and the TELeR taxonomy for controlling prompt detail.
Early design choices were informed by a small teacher-assisted case study, while our final evaluation analyzed over \textbf{1,100} questions from \textbf{19} open-source LLMs. With goal and lab understanding grounding questions in teacher intent and simulation context, the question taxonomy elevates cognitive demand (open-ended formats and relational types raise quality by \textbf{0.29–0.39} points), and optimized TELeR prompts enhance format adherence (\textbf{80\%} parsability, >\textbf{90\%} adherence). Larger models yield the strongest gains: parsability +\textbf{37.1\%}, adherence +\textbf{25.7\%}, and average quality +\textbf{0.8} Likert points.

\end{abstract}

\section{Introduction}\label{sec:intro}




Virtual simulations can enhance science education by enabling hands-on experiments, accessible inquiry-based exploration \cite{puntambekartechnology}, and deeper conceptual understanding \cite{Smetana01062012}. However, without structured guidance, students may struggle to use these tools effectively \cite{falloon2019using}. Third-party instructional materials often misalign with teachers’ specific goals or classroom context, and adapting them—such as creating simulation-specific worksheets—can be time-consuming, fragmented, and impractical \cite{davis2016adaptive, ben2022science}. Targeted support is therefore needed to effectively integrate online simulations into the classroom while ensuring alignment with curriculum objectives and student needs \cite{fischer2020preparing}.



Recent advances in Large Language Models (LLMs) offer a promising avenue for addressing these challenges. With their ability to interpret and generate natural language, LLMs increasingly support high-level educational tasks such as content creation \cite{attard-etal-2024-classgeneration, kehui-etal-2025-classgeneration} and curriculum adaptation \cite{weijers-etal-2024-quantifying, zibo-etal-2024-adaptation}. Notably, they can automate the generation of pedagogically relevant questions \cite{Razafinirina_Dimbisoa_Mahatody_2024}, reducing teacher workload while maintaining instructional depth. These developments highlight the potential of LLMs to act as adaptable co-designers in educational toolchains, a role this work explores in the context of interactive, simulation-based learning.




In this paper, we introduce a novel framework for \textit{instructional goal-aligned question generation}, enabling LLMs to produce pedagogically relevant questions grounded in interactive virtual simulations. Developed with K-12 educators, the framework integrates four key components: (1) instructional goal understanding via teacher–LLM dialogue, (2) lab understanding via knowledge unit and relationship analysis, (3) a question taxonomy to structure cognitive and pedagogical intent, and (4) the TELeR taxonomy to control prompt detail. Informed by a preliminary case study, we conducted an automated evaluation of over \textbf{1,100} questions from \textbf{19} LLMs, assessing structural validity, linguistic clarity, and pedagogical alignment with simulation-based learning goals. Through this, we provide new insights into leveraging LLMs for structured, classroom-ready question generation. Our key contributions are as follows:


\begin{enumerate}[leftmargin=*,itemsep=0ex,partopsep=0ex,parsep=0ex]
    \item We introduce a novel \emph{instructional goal-aligned framework} for question generation in virtual labs, comprising four components: goal understanding, lab understanding, a \textit{newly-introduced} question taxonomy (Section \ref{sec:methods-question}), and the TELeR taxonomy \cite{teler}—to ground questions in instructional intent and structure cognitive demand.
    \item We evaluate the proposed framework on over \textbf{1,100} questions generated across \textbf{19} lightweight, deployment-friendly LLMs to assess their ability to generate structured, contextually relevant questions aligned with classroom learning goals.
    \item We show that larger models yield \textbf{37.1\%} higher structural validity and \textbf{0.8}-point gains in quality, while optimized TELeR prompts enhance format adherence (\textbf{80\%} parsability, >\textbf{90\%} adherence). Quality varies most across question format and type: open-ended formats (free response) and relational types (justification) foster higher-order thinking, improving question quality by \textbf{0.29–0.39} points.
\end{enumerate}
\begin{figure*}[ht]
    \centering
    \includegraphics[scale = 0.45]{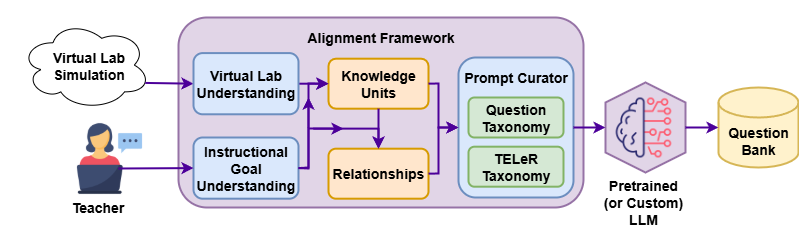}
    \caption{Proposed prompt-based, teacher goal-aligned question generation framework. As input, teachers provide both their instructional goals and a virtual simulation to facilitate student experiments.}
    \label{fig:architecture}
\end{figure*}
\vspace{-2pt}
\section{Related Work}\label{sec:related}

Recent advances in generative AI have opened new opportunities for supporting teachers in instructional planning, assessment design, and classroom adaptation \cite{kwid2024review}. High-capacity LLMs such as GPT-4 have been positioned as general-purpose assistants for generating lesson plans and formative assessments tailored to specific educational contexts \cite{cooper2023examining}. However, teachers often struggle to prompt these models effectively, frequently relying on zero-shot interactions and investing significant time validating and revising outputs \cite{dilling2024using, kuusemets2024assessing}.

To address these challenges, educational AI platforms have begun offering more structured support \cite{niu2022teachers, eden2024review}. For example, Khanmigo provides lesson-planning and content-generation tools \cite{khanmigo}, though these often follow rigid templates \cite{sawyer2024counterexamples, walkington2025implications}. MagicSchool offers broader customization, such as generating questions from YouTube videos \cite{foster2024magicschool}, yet few systems align with specific instructional goals or student experiences in third-party simulations \cite{dai2022educational}. Despite these structured tools, most platforms remain limited in generating questions that are both pedagogically aligned and responsive to the dynamic content of virtual interactive simulations, highlighting the need for more targeted, flexible approaches.

One promising direction that addresses this gap is automated question generation. Prior work explores LLM-assisted generation of learner-adaptive, context-aware questions \cite{Hsiao_Chung_2022, guo-etal-2022-dsm}, including personalized feedback \cite{das2021automatic}, curriculum-aligned assessment \cite{zhang-etal-2022-meta, song2023automatic}, and retrieval-augmented generation \cite{pan2024rag}. While other studies focus on educational question generation \cite{Lee_Jung_Jeon_Sohn_Hwang_Moon_Kim_2023, MAITY2025100370}, few target the exploratory nature of virtual simulations \cite{kurdi2020systematic}, especially third-party labs integrated into teacher-driven lessons, limiting alignment with both interactive context and instructional goals.

Building on these advances in automated question generation, recent work has explored aligning LLMs with human goals, preferences, and values across a range of domains, including education \cite{Razafinirina_Dimbisoa_Mahatody_2024, sonkar-etal-2024-pedagogical}. Within this landscape, aligning model outputs to the specific instructional goals of teachers remains an open challenge in dynamic or exploratory settings like simulation-based learning \cite{simpal}. Our work builds on this by addressing the dual challenge of aligning with both teacher-defined instructional goals and the context of the simulation itself, offering a flexible framework for generating contextually relevant questions in classroom practice.
\section{The Alignment Framework}\label{sec:methods}


In this section, we describe our proposed \textit{prompt-based, teacher goal-aligned question generation framework}, which is designed to ensure that questions generated by LLMs remain closely tied to teachers' instructional goals and the structure of the simulation. At a high level, this framework consists of two interconnected stages: 1) eliciting and representing teacher intent, and 2) generating questions that are aligned with both that intent and the virtual simulation's context. Figure \ref{fig:architecture} depicts this process, showing how the virtual simulation context and teacher-provided goals are transformed into semi-structured representations that guide the generation of pedagogically relevant questions.

\subsection{Virtual Lab Understanding}

To ground question generation in the virtual lab, we first construct a structured representation of its key concepts and interactions. Each simulation is expressed as a semi-structured \textit{simulation representation}, $S_i$, which encodes the knowledge being emphasized and the relationships that organize it:

\vspace{-4mm}
\begin{equation}
    S_i = \left\{
        \begin{array}{ll}
            instruction\_goals: & \texttt{<string>}, \\
            knowledge\_units: & [K_1, K_2, \dots], \\
            relationships: & [R_1, R_2, \dots]
        \end{array}
    \right\}
\end{equation}

In this structure, \textit{Knowledge Units} capture the concepts, variables, or skills that the teacher highlights as central to the lesson. Each knowledge unit records its name, a brief description (with any shorthand), and its type—whether it functions as an input, output, constant, or observable value within the simulation environment.

\textit{Relationships}, in turn, link two or more knowledge units to express conceptual groupings, dependencies, or causal links. Each relationship contains a label, a short description, and the set of knowledge units it connects. For example, a relationship may encode that changing an input variable like \textit{temperature} affects an output variable, \textit{pressure}, or that two observable values belong to the same conceptual category.

This structured representation serves as the pedagogical backbone for question generation, ensuring that questions remain grounded in the simulation's conceptual and interactive structure.

\subsection{Instructional Goal Understanding}\label{sec:methods-understanding}

Jointly, we capture the teacher's instructional intent with a structured dialogue designed to elicit high-level learning objectives, target concepts, and student expectations. Teachers are guided with flexible, open-ended prompts such as:

\vspace{-2mm}
\begin{itemize}[leftmargin=*,itemsep=0ex,partopsep=0ex,parsep=0ex]
    \item \emph{What are the key concepts or phenomena you want students to explore in this simulation?}
   \item \emph{What prior knowledge do students bring into the activity?}
    \item \emph{What kinds of reasoning or analysis should students practice?}
\end{itemize}

These responses are then mapped into the simulation representation, $S_i$, linking the teacher's instructional goals to the underlying knowledge units and relationships. By independently capturing goal understanding and lab structure, the framework ensures that question generation reflects both teacher intent and the simulation's conceptual design.

\subsection{Prompt Curation for Controlled Question Generation}\label{sec:methods-generation}

Once the lab's structure and teacher's instructional intent has been translated into a simulation representation, the framework proceeds to generate candidate questions that engage students with the highlighted concepts and relationships. The simulation representation $S_i$ (Section \ref{sec:methods-understanding}) provides the grounding context, ensuring that generated questions remain tied to teacher-defined goals rather than drifting into generic or irrelevant phrasing.

\subsubsection{Question Taxonomy}\label{sec:methods-question}
To create a diverse pool of prompts, we employ an LLM prompted with subsets of $S_i$. Candidate questions are then produced in several pedagogically useful \textit{question formats}: multiple choice, multiple select, true/false, fill-in-the-blank, and free-response essay—each emphasizing different modes of reasoning and recall.

However, preliminary analysis from our case study (Section \ref{sec:setup-case-study}) revealed that providing the model with \textit{all} available knowledge units and relationships often produced questions that were overly generic or unfocused. To address this, we introduce a targeted alignment step that narrows the prompt scope using a fixed taxonomy of \textit{seven} question types:

\vspace{-1mm}
\begin{itemize}[leftmargin=*,itemsep=0ex,partopsep=0ex,parsep=0ex]
    \item \textbf{Conceptual}: Define or describe a single knowledge unit.
    \item \textbf{Cause-and-Effect}: Explain how one knowledge unit influences another.
    \item \textbf{Critical Thinking/Application}: Apply concepts to a novel or complex scenario.
    \item \textbf{Relationship}: Identify or reason about a specific relationship.
    \item \textbf{Causal Chain}: Trace a sequence of linked changes or effects across multiple knowledge units.
    \item \textbf{Calculation}: Solve quantitative problems using known relationships.
    \item \textbf{Justification}: Argue for or against a particular interpretation of a relationship based on the simulation.
\end{itemize}

Each \textit{question type} is associated with a targeted subset of the simulation representation. For instance, a \textit{conceptual} question may only reference a single knowledge unit, while a \textit{causal chain} spans a path of linked knowledge units within a relationship. This selective conditioning improves both alignment with the teacher's intent and the clarity of the generated output. By combining structured prompt engineering with type-specific input selection, the framework balances variety and depth, producing question sets that are both pedagogically grounded and closely aligned with teacher-defined goals in the simulation context.

\subsubsection{TELeR Taxonomy}
Since large language models exhibit variability in their outputs, we guide the generation process through a multi-level prompt engineering scheme informed by the TELeR taxonomy \cite{teler}. We employ levels 1-4: level 1 provides minimal guidance, level 2 adds more detail, level 3 includes a structured list of considerations, and level 4 specifies the characteristics of an ideal response. This allows us to examine how different degrees of instructional context shape question quality and relevance, balancing breadth of exploration with pedagogical coherence.

\section{Case Study: Initial Teacher Evaluation}\label{sec:setup-case-study}

To inform the design and refinement of our framework, we conducted an initial case study with four K-12 teachers interacting with the system across multiple simulation activities. Participants were tasked with defining knowledge units and relationships, generating questions, and reviewing outputs for clarity and relevance. More details on the context and participants are available in Section \ref{sec:appendix:case_study}.

Overall, teachers found the framework intuitive and straightforward to navigate, particularly when defining knowledge units and relationships. However, despite positive perceptions, technical reliability proved a significant barrier: 68\% of question generation attempts resulted in system-level errors, highlighting the deeper need for system-level validation and robustness improvements.

Teachers also reported that while the generated questions were topically relevant, they often lacked specificity and precise alignment with simulation-based objectives. As a result, the outputs required additional editing and filtering, which reduced the perceived efficiency benefits of the tool.

These observations motivated two key refinements. First, improvements to system reliability were prioritized to reduce errors in question generation. Second, to better control specificity and alignment, we developed a seven-category taxonomy of question types (Section \ref{sec:methods-generation}), allowing generated questions to focus on targeted subsets of knowledge units and relationships. Together, these refinements laid the foundation for the large-scale automated evaluation described in the following sections.

\section{Experimental Setup}\label{sec:experiments}

In this section, we describe the experimental setup, including the curated simulation dataset, the open-source LLMs under study, and the procedures for evaluating structural validity and pedagogical quality, our two key metrics for assessing question generation performance.

\subsection{Dataset}

To evaluate the framework under realistic classroom conditions, we compiled a dataset of eight teacher-system conversations drawn from five distinct simulations. Six of these conversations were reused from the preliminary case study to better reflect authentic, teacher-driven interactions.

For each conversation, the system generated seven distinct question types across five output formats (see Section \ref{sec:methods-generation}), yielding \textbf{35} questions per conversation. Across all conversations, this results in \textbf{280} generated questions per prompt level.

To assess the impact of prompt specificity, we repeated this process using TELeR levels 1 through 4, producing a total of \textbf{1120} questions per model under evaluation.

\subsection{Models Examined}

To align with privacy and security considerations, our evaluation is restricted to lightweight, consumer-accessible, open-source language models. This reflects realistic deployment conditions in educational settings, where data sensitivity and infrastructure constraints are critical.

In total, we evaluate 19 models spanning nine model families: Phi 3.5 \cite{abdin2024phi3technicalreporthighly}, Qwen 2.5 \cite{qwen2.5}, DeepSeek-R1 \cite{deepseekai2025deepseekr1incentivizingreasoningcapability}, Cogito v1 (Preview), LLaMA 3 \cite{grattafiori2024llama3herdmodels}, Falcon 3 \cite{Falcon3}, OLMo 2 \cite{olmo20252olmo2furious}, Ministral \cite{jiang2023mistral7b}, and Gemma 2 \& 3 \cite{gemmateam2024gemma2improvingopen, gemmateam2025gemma3technicalreport}.

\subsection{Model Parameters}

All models were run with consistent generation parameters to ensure fair comparison. Specifically, the sampling parameters, $top\_k$ and $top\_p$, were fixed at their default values of $-1$ and $1$, respectively \cite{kwon2023efficient-vllm}, ensuring that all tokens were considered during generation. The inference temperature was held constant at $0.2$ to promote deterministic outputs while still allowing minor variations in phrasing and structure.

\subsection{Metrics}

To evaluate model performance, we assess both \emph{validity}, which captures structural correctness of outputs, and \emph{quality}, which reflects the pedagogical and linguistic value of the generated questions.

\subsubsection{Validity}\label{sec:experiments-validity}

Structured generation remains a known challenge for LLMs, even with explicit guidance. Our validity evaluation measures how reliably models produce outputs that adhere to the expected question formats across different models, question types, and prompt levels. We quantify validity using three complementary metrics:

\vspace{-1mm}
\begin{itemize}[leftmargin=*,itemsep=0ex,partopsep=0ex,parsep=0ex]
    \item \textbf{JSON load accuracy}: The proportion of generations that successfully parse as valid JSON objects.
    \item \textbf{Format accuracy}: The proportion of outputs that conform to the expected question format (e.g., a fill-in-the-blank question contains an explicit blank).
    \item \textbf{Existence accuracy}: A coarse-grained structural score (up to 5 points), assigning one point per expected question format. For example, if a model generates valid questions in all formats except the free-response essay, it receives a score of 4.
\end{itemize}

Together, these metrics capture whether models consistently produce well-formed and structurally valid outputs under varying prompt constraints.

\subsubsection{Quality}\label{sec:experiments-quality}

Beyond structural correctness, we assess the pedagogical/linguistic quality of generated questions, focusing on alignment with instructional goals and practical utility in classroom settings. Each question is rated on a 5-point Likert scale (1 = ``absolutely not'', 5 = ``yes, definitely'') across ten criteria:

\begin{enumerate}[leftmargin=*,itemsep=0ex,partopsep=0ex,parsep=0ex]
    \item \textbf{Fluency}: Is the question grammatically fluent and natural-sounding?
    \item \textbf{Correctness}: Is all presented information accurate and factually correct?
    \item \textbf{Clarity}: Is the question easy to interpret and understand?
    \item \textbf{Specificity}: Is the question sufficiently precise to elicit a focused response?
    \item \textbf{Bias}: Is the question free from leading language or bias?
    \item \textbf{Relevance}: Is the question directly related to the simulation and learning objectives?
    \item \textbf{Practicality}: Would this question be useful in a real instructional context?
    \item \textbf{Alignment}: Does the question align with the established goals or expectations of the lab?
    \item \textbf{Feasibility}: Can the question be answered based on information available in the simulation?
    \item \textbf{Critical Thinking}: Does the question encourage deeper reasoning or reflection?
\end{enumerate}

To ensure robustness and reduce evaluator bias, each question is independently rated by three high-capacity models—GPT 4.1 \cite{openai2024gpt4technicalreport}, Gemini 2.0 \cite{geminiteam2024geminifamilyhighlycapable}, and Claude 3.7 \cite{Claude3S}—using a standardized rubric. Ratings are aggregated to produce a composite quality score per question, with Krippendorff's alpha ($\alpha_k$) computed to assess inter-rater consistency \cite{hayes2007answering}.
\section{Results}\label{sec:results}

\subsection{Validity Results}

As described in section \ref{sec:experiments-validity}, we assess validity using three core metrics—JSON load accuracy, format accuracy, and existence accuracy—evaluated across models, prompt levels, question formats, and question types. Tables \ref{tbl:validity}, \ref{tbl:validity_teler}, \ref{tbl:validity_format}, and \ref{tbl:validity_type} summarize performance across each of these dimensions.

\noindent\textbf{Size Validity}: Table \ref{tbl:validity} reveals clear patterns across the validity metrics. JSON accuracy is generally strong (>80\%) for models with $\geq$3B parameters and increases with model size by roughly 37.1\%, though gains taper beyond 3B. Format adherence is also high (>85\%) for models $\geq$3B parameters, improving by about 25.7\% with increasing size. In contrast, existence scores show limited correlation with model size. Most models exceed 4 points, indicating strong coverage of all expected question types, though LLaMA-3.1-8B and the DeepSeek family score notably lower.

\begin{table}[!b] \small
    \centering
    \resizebox{\columnwidth}{!}{
    \begin{tabular}{c||c|c|c}
        \hline
         Model
         & \begin{tabular}[c]{@{}c@{}} JSON\end{tabular}
         & \begin{tabular}[c]{@{}c@{}} Format\end{tabular}
         & \begin{tabular}[c]{@{}c@{}} Existence\end{tabular}\\
         \hline\hline
Phi-3.5-mini		& 0.888	& 0.986	& 4.246	\\
\hline
LLaMA-3.1-8B		& 0.228	& \highlight\textbf{1.000}	& 1.000	\\
\hline
LLaMA-3.2-1B		& 0.549	& 0.566	& 4.472	\\
\hline
LLaMA-3.2-3B		& 0.920	& 0.823	& 4.515	\\
\hline
Qwen-2.5-0.5B		& 0.420	& 0.719	& 4.309	\\
\hline
Qwen-2.5-1.5B		& \highlight\textbf{1.000}	& 0.775	& 4.853	\\
\hline
Qwen-2.5-3B		& 0.920	& 0.934	& 4.112	\\
\hline
Qwen-2.5-7B		& 0.862	& 0.988	& \highlight\textbf{4.990}	\\
\hline
Falcon-3-3B		& 0.875	& 0.888	& 4.923	\\
\hline
Falcon-3-7B		& 0.804	& 0.982	& 4.061	\\
\hline
OLMo-2-7B		& 0.714	& 0.800	& 4.081	\\
\hline
Ministral-8B		& \highlight\textbf{1.000}	& 0.968	& 4.705	\\
\hline
Gemma-2-2B		& 0.978	& 0.718	& 4.808	\\
\hline
Gemma-3-4B		& 0.933	& 0.937	& 4.584	\\
\hline
DeepSeek-Q-1.5B		& 0.750	& 0.860	& 3.185	\\
\hline
DeepSeek-Q-7B		& 0.732	& 0.898	& 3.902	\\
\hline
DeepSeek-L-8B		& 0.656	& 0.984	& 3.599	\\
\hline
Cogito-v1-L-3B		& 0.790	& 0.831	& 4.531	\\
\hline
Cogito-v1-L-8B		& 0.862	& 0.981	& 4.637	\\
\hline
    \end{tabular}
    }
    \caption{Validity results for each model tested. Top performers are highlighted and marked in \textbf{bold}. Models suffixed with ``-Q'' are Qwen-distilled models, and Models suffixed with ``-L'' are LLaMA-distilled models. The Cogito v1 models are preview models.}
    \label{tbl:validity}
\end{table}
\begin{table}[!htb] \small
    \centering
    
    \resizebox{0.9\columnwidth}{!}{
    \begin{tabular}{c||c|c|c}
        \hline
         TELeR Level
         & \begin{tabular}[c]{@{}c@{}} JSON\end{tabular}
         & \begin{tabular}[c]{@{}c@{}} Format\end{tabular}
         & \begin{tabular}[c]{@{}c@{}} Existence\end{tabular}\\
         \hline\hline
TELeR-L1		& 0.796	& 0.905	& 4.484	\\
\hline
TELeR-L2		& \highlight\textbf{0.800}	& 0.900	& 4.462	\\
\hline
TELeR-L3		& 0.767	& \highlight\textbf{0.913}	& 4.482	\\
\hline
TELeR-L4		& 0.770	& 0.905	& \highlight\textbf{4.530}	\\
\hline
    \end{tabular}
    }
    \caption{Validity results for each TELeR level tested. Top performers are highlighted and marked in \highlight\textbf{bold}.}
    \label{tbl:validity_teler}
\end{table}
\begin{table}[!htb] \small
    \centering
    
    \resizebox{0.9\columnwidth}{!}{
    \begin{tabular}{c||c|c}
        \hline
         Question Format
         & \begin{tabular}[c]{@{}c@{}} Format\end{tabular}
         & \begin{tabular}[c]{@{}c@{}} Existence\end{tabular}\\
         \hline\hline
multiple choice		& 0.864	& \highlight\textbf{0.967}	\\
\hline
multiple select		& 0.992	& 0.891	\\
\hline
true/false		& \highlight\textbf{0.999}	& 0.940	\\
\hline
fill in the blank		& 0.679	& 0.936	\\
\hline
free response essay		& 0.996	& 0.756	\\
\hline
    \end{tabular}
    }

    \caption{Validity results for each question format tested. Top performers are highlighted and marked in \highlight\textbf{bold}.}
    \label{tbl:validity_format}
\end{table}
\begin{table}[!htb] \small
    \centering
    
    \resizebox{\columnwidth}{!}{
    \begin{tabular}{c||c|c|c}
        \hline
         Question Type
         & \begin{tabular}[c]{@{}c@{}} JSON\end{tabular}
         & \begin{tabular}[c]{@{}c@{}} Format\end{tabular}
         & \begin{tabular}[c]{@{}c@{}} Existence\end{tabular}\\
         \hline\hline
conceptual		& \highlight\textbf{0.831}	& 0.908	& 4.418	\\
\hline
cause-and-effect		& 0.817	& 0.896	& \highlight\textbf{4.648}	\\
\hline
critical thinking		& 0.786	& 0.905	& 4.470	\\
\hline
relationship		& 0.819	& 0.889	& 4.556	\\
\hline
causal chain		& 0.729	& 0.915	& 4.594	\\
\hline
calculation		& 0.794	& 0.903	& 4.408	\\
\hline
justification		& 0.706	& \highlight\textbf{0.924}	& 4.334	\\
\hline
    \end{tabular}
    }

    \caption{Validity results for each question type tested. Top performers are highlighted and marked in \highlight\textbf{bold}.}
    \label{tbl:validity_type}
\end{table}

\noindent\textbf{Detail Validity}: Table \ref{tbl:validity_teler} shows that prompt detail has a modest effect on validity metrics. JSON accuracy remains consistently between 76–80\%, with Level 2 prompts performing best at 80\%, suggesting that additional detail beyond Level 2 may distract the model from producing syntactically correct outputs. Format adherence is very strong across all levels (>90\%), with Level 3 prompts achieving the highest performance at 91.3\%. Existence scores hold steady between 4.4–4.5, indicating that all question types are reliably generated regardless of prompt specificity.

\begin{table*}[ht]
    \centering
    \resizebox{\textwidth}{!}{
    \begin{tabular}{c||c|c|c|c|c|c|c|c|c|c|c|c}
        \hline
         Model
         & \begin{tabular}[c]{@{}c@{}} Fluency\end{tabular}
         & \begin{tabular}[c]{@{}c@{}} Correctness\end{tabular}
         & \begin{tabular}[c]{@{}c@{}} Clarity\end{tabular}
         & \begin{tabular}[c]{@{}c@{}} Specificity\end{tabular}
         & \begin{tabular}[c]{@{}c@{}} Bias\end{tabular}
         & \begin{tabular}[c]{@{}c@{}} Relevance\end{tabular}
         & \begin{tabular}[c]{@{}c@{}} Practicality\end{tabular}
         & \begin{tabular}[c]{@{}c@{}} Alignment\end{tabular}
         & \begin{tabular}[c]{@{}c@{}} Feasibility\end{tabular}
         & \begin{tabular}[c]{@{}c@{}} Critical\end{tabular}
         & \begin{tabular}[c]{@{}c@{}} Average\end{tabular}
         & \begin{tabular}[c]{@{}c@{}} $\alpha_k$\end{tabular}\\
         \hline\hline
Phi-3.5-mini		& 4.866	& 4.642	& 4.684	& 4.708	& 4.973	& 4.576	& 4.354	& 4.329	& 4.296	& 3.368	& 4.480	& 0.661	\\
\hline
LLaMA-3.1-8B		& 4.869	& \highlight\textbf{4.745}	& 4.627	& 4.359	& 4.908	& 4.431	& 4.405	& 4.190	& 3.961	& \highlight\textbf{4.451}	& 4.495	& 0.617	\\
\hline
LLaMA-3.2-1B		& 4.253	& 3.261	& 3.773	& 3.776	& 4.580	& 3.994	& 3.353	& 3.522	& 3.483	& 2.724	& 3.672	& 0.631	\\
\hline
LLaMA-3.2-3B		& 4.733	& 4.131	& 4.404	& 4.303	& 4.901	& 4.301	& 3.982	& 3.916	& 3.893	& 3.254	& 4.182	& 0.656	\\
\hline
Qwen-2.5-0.5B		& 4.092	& 3.635	& 3.675	& 3.560	& 4.676	& 4.067	& 3.468	& 3.635	& 3.481	& 2.691	& 3.698	& 0.691	\\
\hline
Qwen-2.5-1.5B		& 4.822	& 4.188	& 4.548	& 4.422	& 4.891	& 4.491	& 4.139	& 4.168	& 4.114	& 3.175	& 4.296	& 0.673	\\
\hline
Qwen-2.5-3B		& 4.745	& 4.287	& 4.459	& 4.510	& 4.910	& 4.521	& 4.177	& 4.221	& 4.233	& 3.067	& 4.313	& 0.636	\\
\hline
Qwen-2.5-7B		& 4.882	& 4.632	& 4.714	& 4.662	& 4.951	& 4.621	& 4.435	& 4.408	& 4.425	& 3.460	& \highlight\textbf{4.519}	& 0.702	\\
\hline
Falcon-3-3B		& 4.780	& 4.485	& 4.542	& 4.510	& 4.920	& 4.459	& 4.145	& 4.150	& 4.135	& 3.207	& 4.333	& 0.654	\\
\hline
Falcon-3-7B		& 4.825	& 4.608	& 4.649	& 4.654	& 4.955	& 4.662	& 4.345	& 4.423	& \highlight\textbf{4.448}	& 3.100	& 4.467	& 0.681	\\
\hline
OLMo-2-7B		& 4.740	& 4.426	& 4.437	& 4.478	& 4.934	& \highlight\textbf{4.690}	& 4.354	& 4.423	& 4.355	& 3.501	& 4.434	& 0.592	\\
\hline
Ministral-8B		& 4.914	& 4.450	& \highlight\textbf{4.725}	& 4.594	& 4.946	& 4.587	& 4.289	& 4.286	& 4.343	& 3.160	& 4.429	& 0.658	\\
\hline
Gemma-2-2B		& \highlight\textbf{4.917}	& 4.263	& 4.694	& 4.480	& 4.930	& 4.529	& 4.272	& 4.226	& 4.161	& 3.509	& 4.398	& 0.649	\\
\hline
Gemma-3-4B		& 4.910	& 4.451	& 4.723	& \highlight\textbf{4.757}	& \highlight\textbf{4.984}	& 4.570	& 4.324	& 4.264	& 4.310	& 3.371	& 4.466	& 0.663	\\
\hline
DeepSeek-Q-1.5B		& 4.532	& 4.179	& 4.249	& 4.206	& 4.782	& 4.384	& 3.970	& 4.070	& 4.108	& 2.883	& 4.136	& 0.702	\\
\hline
DeepSeek-Q-7B		& 4.882	& 4.581	& 4.673	& 4.548	& 4.954	& 4.688	& \highlight\textbf{4.449}	& \highlight\textbf{4.441}	& 4.418	& 3.356	& 4.499	& 0.646	\\
\hline
DeepSeek-L-8B		& 4.730	& 4.355	& 4.475	& 4.354	& 4.928	& 4.516	& 4.166	& 4.208	& 4.226	& 3.123	& 4.308	& 0.670	\\
\hline
Cogito-v1-L-3B		& 4.782	& 4.257	& 4.507	& 4.329	& 4.900	& 4.367	& 4.029	& 4.011	& 3.959	& 3.209	& 4.235	& 0.657	\\
\hline
Cogito-v1-L-8B		& 4.838	& 4.477	& 4.618	& 4.498	& 4.933	& 4.621	& 4.309	& 4.341	& 4.328	& 3.275	& 4.424	& 0.679	\\
\hline
    \end{tabular}
    }
    \caption{Quality results for each model tested.. Top performers are highlighted and marked in \highlight\textbf{bold}. Models suffixed with ``-Q'' are Qwen-distilled models, and Models suffixed with ``-L'' are LLaMA-distilled models. The Cogito v1 models are preview models.}
    \label{tbl:quality}
\end{table*}
\begin{table*}
    \centering
    \resizebox{\textwidth}{!}{
    \begin{tabular}{c||c|c|c|c|c|c|c|c|c|c|c|c}
        \hline
         TELeR Level
         & \begin{tabular}[c]{@{}c@{}} Fluency\end{tabular}
         & \begin{tabular}[c]{@{}c@{}} Correctness\end{tabular}
         & \begin{tabular}[c]{@{}c@{}} Clarity\end{tabular}
         & \begin{tabular}[c]{@{}c@{}} Specificity\end{tabular}
         & \begin{tabular}[c]{@{}c@{}} Bias\end{tabular}
         & \begin{tabular}[c]{@{}c@{}} Relevance\end{tabular}
         & \begin{tabular}[c]{@{}c@{}} Practicality\end{tabular}
         & \begin{tabular}[c]{@{}c@{}} Alignment\end{tabular}
         & \begin{tabular}[c]{@{}c@{}} Feasibility\end{tabular}
         & \begin{tabular}[c]{@{}c@{}} Critical\end{tabular}
         & \begin{tabular}[c]{@{}c@{}} Average\end{tabular}
         & \begin{tabular}[c]{@{}c@{}} $\alpha_k$\end{tabular}\\
         \hline\hline
TELeR-L1		& 4.781	& 4.378	& 4.549	& 4.486	& \highlight\textbf{4.927}	& 4.517	& 4.194	& 4.206	& 4.183	& 3.192	& 4.341	& 0.687	\\
\hline
TELeR-L2		& 4.794	& 4.366	& 4.550	& 4.480	& 4.921	& \highlight\textbf{4.526}	& \highlight\textbf{4.214}	& 4.222	& 4.208	& 3.235	& 4.352	& 0.665	\\
\hline
TELeR-L3		& 4.792	& \highlight\textbf{4.384}	& \highlight\textbf{4.556}	& 4.475	& 4.906	& 4.512	& 4.209	& \highlight\textbf{4.222}	& \highlight\textbf{4.227}	& 3.254	& \highlight\textbf{4.354}	& 0.669	\\
\hline
TELeR-L4		& \highlight\textbf{4.803}	& 4.353	& 4.550	& \highlight\textbf{4.491}	& 4.910	& 4.510	& 4.210	& 4.217	& 4.205	& \highlight\textbf{3.282}	& 4.353	& 0.669	\\
\hline
    \end{tabular}
    }
    \caption{Quality results for each TELeR level tested. Top performers are highlighted and marked in \highlight\textbf{bold}.}
    \label{tbl:quality_teler}
\end{table*}

\noindent\textbf{Format Validity}: Table \ref{tbl:validity_format} summarizes format-level performance across question types. Note that JSON scores are not reported, since successful parsing is a prerequisite for evaluating format correctness. Straightforward formats—multiple select, true/false, and free response—achieve near-perfect format accuracy (>99\%) and strong existence scores (89–94\%). In contrast, more complex structures such as multiple choice and fill-in-the-blank are more error-prone: multiple choice (86.4\% adherence) often incorrectly includes multiple correct answers (mimicking multiple select), while fill-in-the-blank (67.9\% adherence) frequently omits the required blanks.

\noindent\textbf{Type Validity}: Table \ref{tbl:validity_type} presents validity metrics broken down by question type. JSON accuracy ranges from 70–83\%, with conceptual, relationship, and cause-and-effect questions performing at the higher end (>80\%), possibly reflecting their focus on a limited number of knowledge units. Format and existence scores remain consistently high (>88\% and >4.3, respectively), indicating strong model compliance once JSON parsing succeeds.

Overall, these results demonstrate that while structural validity can be achieved across a range of settings, model choice, prompt design, and output format strongly shape reliability. Next, we turn to quality-focused results to examine how these structural patterns translate into pedagogical value.

\subsection{Quality Results}

\begin{table*}[t]
    \centering
    \resizebox{\textwidth}{!}{
    \begin{tabular}{c||c|c|c|c|c|c|c|c|c|c|c|c}
        \hline
         Question Format
         & \begin{tabular}[c]{@{}c@{}} Fluency\end{tabular}
         & \begin{tabular}[c]{@{}c@{}} Correctness\end{tabular}
         & \begin{tabular}[c]{@{}c@{}} Clarity\end{tabular}
         & \begin{tabular}[c]{@{}c@{}} Specificity\end{tabular}
         & \begin{tabular}[c]{@{}c@{}} Bias\end{tabular}
         & \begin{tabular}[c]{@{}c@{}} Relevance\end{tabular}
         & \begin{tabular}[c]{@{}c@{}} Practicality\end{tabular}
         & \begin{tabular}[c]{@{}c@{}} Alignment\end{tabular}
         & \begin{tabular}[c]{@{}c@{}} Feasibility\end{tabular}
         & \begin{tabular}[c]{@{}c@{}} Critical\end{tabular}
         & \begin{tabular}[c]{@{}c@{}} Average\end{tabular}
         & \begin{tabular}[c]{@{}c@{}} $\alpha_k$\end{tabular}\\
         \hline\hline
multiple choice		& \highlight\textbf{4.887}	& 4.560	& \highlight\textbf{4.699}	& \highlight\textbf{4.752}	& \highlight\textbf{4.967}	& 4.529	& 4.363	& 4.250	& \highlight\textbf{4.272}	& 3.215	& 4.449	& 0.679	\\
\hline
multiple select		& 4.706	& 4.117	& 4.401	& 4.594	& 4.906	& \highlight\textbf{4.567}	& 4.289	& 4.259	& 4.210	& 3.432	& 4.348	& 0.600	\\
\hline
true/false		& 4.880	& 4.237	& 4.629	& 4.463	& 4.864	& 4.467	& 3.911	& 4.145	& 4.195	& 2.718	& 4.251	& 0.672	\\
\hline
fill in the blank		& 4.563	& 4.477	& 4.322	& 4.230	& 4.916	& 4.448	& 4.028	& 4.125	& 4.211	& 2.593	& 4.191	& 0.729	\\
\hline
free response essay		& 4.844	& \highlight\textbf{4.568}	& 4.633	& 4.251	& 4.940	& 4.557	& \highlight\textbf{4.458}	& \highlight\textbf{4.292}	& 4.134	& \highlight\textbf{4.217}	& \highlight\textbf{4.489}	& 0.686	\\
\hline
    \end{tabular}
    }
    \caption{Quality results for each question format tested. Top performers are highlighted and marked in \highlight\textbf{bold}.}
    \label{tbl:quality_format}
\end{table*}
\begin{table*}
    \centering
    \resizebox{\textwidth}{!}{
    \begin{tabular}{c||c|c|c|c|c|c|c|c|c|c|c|c}
        \hline
         Question Type
         & \begin{tabular}[c]{@{}c@{}} Fluency\end{tabular}
         & \begin{tabular}[c]{@{}c@{}} Correctness\end{tabular}
         & \begin{tabular}[c]{@{}c@{}} Clarity\end{tabular}
         & \begin{tabular}[c]{@{}c@{}} Specificity\end{tabular}
         & \begin{tabular}[c]{@{}c@{}} Bias\end{tabular}
         & \begin{tabular}[c]{@{}c@{}} Relevance\end{tabular}
         & \begin{tabular}[c]{@{}c@{}} Practicality\end{tabular}
         & \begin{tabular}[c]{@{}c@{}} Alignment\end{tabular}
         & \begin{tabular}[c]{@{}c@{}} Feasibility\end{tabular}
         & \begin{tabular}[c]{@{}c@{}} Critical\end{tabular}
         & \begin{tabular}[c]{@{}c@{}} Average\end{tabular}
         & \begin{tabular}[c]{@{}c@{}} $\alpha_k$\end{tabular}\\
         \hline\hline
conceptual		& \highlight\textbf{4.859}	& 4.435	& \highlight\textbf{4.670}	& 4.555	& \highlight\textbf{4.954}	& 4.248	& 3.969	& 3.873	& 3.893	& 2.963	& 4.242	& 0.742	\\
\hline
cause-and-effect		& 4.674	& 4.334	& 4.408	& 4.386	& 4.890	& 4.711	& 4.342	& 4.455	& 4.357	& 3.376	& 4.393	& 0.621	\\
\hline
critical thinking		& 4.831	& 4.438	& 4.603	& 4.524	& 4.931	& 4.607	& 4.309	& 4.313	& 4.233	& \highlight\textbf{3.435}	& 4.422	& 0.636	\\
\hline
relationship		& 4.768	& 4.203	& 4.449	& 4.361	& 4.861	& 4.121	& 3.860	& 3.699	& 3.925	& 3.052	& 4.130	& 0.650	\\
\hline
causal chain		& 4.786	& 4.343	& 4.544	& 4.464	& 4.911	& 4.621	& 4.287	& 4.362	& 4.279	& 3.272	& 4.387	& 0.692	\\
\hline
calculation		& 4.810	& 4.376	& 4.581	& 4.555	& 4.941	& 4.598	& 4.271	& 4.330	& 4.290	& 3.218	& 4.397	& 0.670	\\
\hline
justification		& 4.833	& \highlight\textbf{4.492}	& 4.626	& \highlight\textbf{4.559}	& 4.930	& \highlight\textbf{4.773}	& \highlight\textbf{4.477}	& \highlight\textbf{4.576}	& \highlight\textbf{4.540}	& 3.410	& \highlight\textbf{4.522}	& 0.635	\\
\hline
    \end{tabular}
    }
    \caption{Quality results for each question type tested. Top performers are highlighted and marked in \highlight\textbf{bold}.}
    \label{tbl:quality_type}
\end{table*}

\noindent\textbf{Size Quality}: Table \ref{tbl:quality} shows a clear positive relationship between model size and question quality. Average scores increase by up to 0.82 points (out of 5) as size grows, suggesting that larger models not only improve structural validity but also enhance linguistic and pedagogical strength. Importantly, even smaller models cluster toward the upper end of the scale (3.67–4.39), indicating broadly reliable quality once valid outputs are achieved.

\noindent\textbf{Detail Quality}: As shown in Table \ref{tbl:quality_teler}, prompt specificity has a modest but consistent effect. Scores generally increase from Levels 1 through 3, peaking at Level 3 before leveling off. This suggests that moderate detail provides the best balance of structure and flexibility, while additional elaboration beyond Level 3 offers limited benefit.


\noindent\textbf{Format Quality}: Table \ref{tbl:quality_format} shows that while overall quality remains high, critical thinking scores reveal a clear hierarchy: fill-in-the-blank and true/false score lowest (2.59–2.72), multiple choice (3.21) and multiple select (3.43) fall in the middle, and free response alone exceeds 4 points (4.21). This trend reflects the increasing cognitive demands of the formats, from recall and recognition to open-ended reasoning. Other dimensions (fluency, clarity, feasibility) remain consistently strong, underscoring critical thinking as the primary differentiator across formats.


\noindent\textbf{Type Quality}: As summarized in Table \ref{tbl:quality_type}, justification questions consistently score highest, with differences of up to 0.39 points on average. Critical thinking-type questions also excel in their targeted category (+0.1 points), indicating that their design reliably supports higher-order reasoning. In contrast, conceptual and relationship-type questions score lowest across several alignment-focused dimensions, making them the weakest overall performers (-0.28–0.39 points). These simpler question types, which probe knowledge units without an explicit relational frame, tend to function more like general recall questions, limiting their simulation specificity and reducing alignment with instructional goals.

Taken together, these findings show that once structural validity is achieved, question quality remains consistently strong across models, prompt levels, formats, and types. The main sources of variation arise in how different formats and types support critical thinking and contextual alignment: open-ended formats (e.g., free response) and relationally grounded types (e.g., justification) promote deeper reasoning and stronger alignment, while simpler forms (e.g., fill-in-the-blank, conceptual) more often default to more general recall.
\section{Conclusion}\label{sec:conclusion}

In this paper, we introduced a novel \textit{prompt-based, instructional goal-aligned framework} for generating contextually aligned, pedagogically meaningful questions from natural language prompts, designed to operate across a wide range of third-party virtual science simulations.

Our evaluation revealed several insights about generation reliability and educational value, underscoring the importance of linking question generation directly to instructional goals and simulation structure. Structural validity remains the primary bottleneck: larger models ($\geq$3B parameters) show \textbf{37.1\%} higher reliablity in producing syntactically correct, JSON-parsable outputs, though size alone is not a guarantee of success. Prompt specificity plays a meaningful role, with moderate detail (TELeR levels 2-3) providing the best balance between syntactic accuracy and linguistic quality.

Once structural validity is achieved, question quality remains consistently high across models (\textbf{4.1–4.5} points), with variation emerging primarily along two dimensions: the cognitive demands of the question format and the relational depth of the question type. Formats that require open-ended responses and types that explicitly encode relationships consistently encourage higher-order thinking and stronger instructional alignment, whereas simpler forms more often default to general recall.


Based on these findings, we propose the following recommendations for deploying LLM-based question generation in classroom contexts:
\begin{enumerate}\setlength\itemsep{0cm}
	\item \textbf{Prioritize structural validity first.} \\
	Ensure that outputs consistently meet JSON and format constraints before optimizing for higher-order quality.
	\item \textbf{Select models strategically.} \\
	Models above 3B parameters generally improve validity and quality, but smaller, more efficient models can still perform competitively when paired with strong prompts—a key consideration for cost-sensitive or privacy-focused environments.
	\item \textbf{Use moderate prompt detail.} \\
	Prompts structured at TELeR Levels 2–3 maximize syntactic reliability without overwhelming the model, striking the best balance between control and flexibility.
	\item \textbf{Match question formats to cognitive goals.} \\
	Use simpler structures (e.g., fill-in-the-blank, true/false) for recall-oriented objectives and more open-ended formats (e.g., free response, multiple select) to target deeper reasoning.
	\item \textbf{Be intentional with question type selection.} \\
	Relationally grounded types (e.g., justification, cause-and-effect) enhance alignment and promote higher-order thinking, while conceptual questions are best used sparingly to support foundational knowledge checks.
\end{enumerate}

Ultimately, effective question generation depends less on raw model size than on the structured integration of goal analysis, simulation understanding, and prompt design. By grounding generation in clear instructional intent, educators and developers can confidently integrate generative models to support inquiry-driven science learning.

\section{Limitations}\label{sec:limitations}

While our findings demonstrate the promise of LLMs in supporting simulation-aligned question generation, several limitations remain. First, our evaluation focused on small and mid-sized open-source models. This choice supports privacy-sensitive educational deployment, but future work should examine how larger models perform under similar conditions.

Second, although our taxonomy of question types improved format adherence, models still occasionally failed to produce fully parsable outputs. More advanced methods, such as schema-aware decoding, may further reduce these errors.

Third, our quality assessments relied on automated LLM-based Likert ratings. While scalable and internally consistent, these methods may not fully capture classroom-specific nuances or reflect the real-world preferences of educators. Incorporating evaluations by practicing educators will be essential to validate instructional relevance.

Finally, although the framework is designed to be simulation-agnostic, our tests used a curated subset of physics-based science simulations. Assessing generalizability across broader domains and content areas remains an important direction for future work.

\section{Ethical Considerations}\label{sec:ethics}

While our work presents relatively low-risk, we acknowledge potential ethical considerations related to the classroom use of LLM-generated content. First, there is a risk that teachers may over-rely on automatically generated content without adequate review, potentially affecting instructional quality. Second, despite efforts to structure and constrain outputs, models may still produce pedagogically inappropriate or scientifically inaccurate questions, which could mislead students if left uncorrected. Finally, although the framework is designed to be accessible to teachers without deep technical expertise, there may still be challenges in ensuring equitable and effective use across diverse educational contexts.
To help address these concerns, our framework emphasizes a human-in-the-loop design that promotes active teacher engagement and oversight, ensuring generated content remains pedagogically relevant and accurate. We encourage future work to continue developing supports that broaden accessibility and strengthen teacher agency in AI-integrated classrooms.



\bibliography{anthology}

\appendix
\section{Appendix}\label{sec:appendix}

\subsection{Case Study: Context and Participants}\label{sec:appendix:case_study}

The data used in this work was collected as part of an IRB-approved case study designed to inform the development of our proposed framework. Four middle school science teachers from two public school districts in the Midwestern United States participated. Three teachers had 15-24 years of experience teaching science, and one had 3 years of experience. Participants were recruited from middle schools within approximately one hour of the university, and only science teachers were included due to the high availability of science-oriented virtual labs. Participation was voluntary and no monetary compensation was provided; however, participating schools were reimbursed for substitute teacher costs.

Informed consent was obtained from all participants prior to data collection. Teachers were fully informed about the purpose of the case study, the types of data being collected, and how their data would inform the development of the framework. Each participant signed a consent form outlining these details, with all collected data being linked to private user IDs and anonymized prior to analysis.

During the case study, two of the system’s developers introduced the question-generation process through a brief demonstration of each step: (1) selecting a virtual lab simulation, (2) describing instructional goals, (3) refining relevant knowledge units and relationships, and (4) generating questions. Teachers then participated in a hands-on activity, first using a provided simulation and later using one of their choice. Multiple testing and feedback rounds were conducted using different LLMs for question generation. Throughout these sessions, researchers provided troubleshooting support and recorded teacher feedback on difficulties, potential issues, suggested improvements, and model preferences based on alignment with instructional goals.

Following the study, teachers completed a post-study questionnaire featuring both Likert-style and open-ended items. The Likert-style questions assessed the perceived usability of system features (rated from \textit{Very difficult} to \textit{Very easy}) and usefulness in achieving learning goals (rated from \textit{Never} to \textit{Always}). Each question included space for written explanations, and open-ended items asked teachers about their overall impressions of the system, likelihood of classroom adoption, and the most and least useful features.

\end{document}